\documentclass[conference]{IEEEtran}
\IEEEoverridecommandlockouts
\usepackage{cite}
\usepackage{amsmath,amssymb,amsfonts}
\usepackage{amsthm}
\usepackage{algorithm}
\usepackage[noend]{algpseudocode}
\usepackage{graphicx}
\usepackage{textcomp}
\usepackage{xcolor}
\usepackage{mathtools}
\usepackage{subcaption}
\usepackage{siunitx}
\usepackage[square, numbers]{natbib}
\usepackage{varwidth}
\usepackage{multirow}

\theoremstyle{definition}

\newcommand{\Loss}{\mathcal{L}}

\DeclarePairedDelimiterX{\infdivx}[2]{(}{)}{%
  #1\delimsize\|#2%
}
\newcommand{\KL}{\displaystyle D_{\text{KL}}\infdivx}
\DeclarePairedDelimiter{\norm}{\lVert}{\rVert}

\newcommand{\series}[1]{\textbf{#1}}
\newcommand{\RNum}[1]{\uppercase\expandafter{\romannumeral #1\relax}}

\DeclareSIUnit{\belmilliwatt}{Bm}
\DeclareSIUnit{\dBm}{\deci\belmilliwatt}

\def\BibTeX{{\rm B\kern-.05em{\sc i\kern-.025em b}\kern-.08em
    T\kern-.1667em\lower.7ex\hbox{E}\kern-.125emX}}
\begin{document}

\title{Automated Antenna Testing Using \\ Encoder-Decoder-based Anomaly Detection}

\author{}
\author{
\IEEEauthorblockN{1\textsuperscript{st} Hans Hao-Hsun Hsu}
\IEEEauthorblockA{
\textit{Huawei Tech. / TU Munich} \\
Munich, Germany \\
hans.hsu@tum.de}
\and
\IEEEauthorblockN{2\textsuperscript{nd} Jiawen Xu}
\IEEEauthorblockA{
\textit{Huawei Technologies} \\
Munich, Germany \\
jiawenxu@huawei.com}
\and
\IEEEauthorblockN{3\textsuperscript{rd} Ravi Sama}
\IEEEauthorblockA{
\textit{Huawei Technologies} \\
Munich, Germany \\
ravi.sama@huawei.com}
\and
\IEEEauthorblockN{4\textsuperscript{th} Matthias Kovatsch}
\IEEEauthorblockA{
\textit{Huawei Technologies} \\
Munich, Germany \\
matthias.kovatsch@huawei.com}
}

\maketitle

\begin{abstract}
We propose a new method for testing antenna arrays that records the radiating electromagnetic (EM) field using an absorbing material and evaluating the resulting thermal image series through an AI using a conditional encoder-decoder model.
Given the power and phase of the signals fed into each array element, we are able to reconstruct normal sequences through our trained model and compare it to the real sequences observed by a thermal camera.
These thermograms only contain low-level patterns such as blobs of various shapes. 
A contour-based anomaly detector can then map the reconstruction error matrix to an anomaly score to identify faulty antenna arrays and increase the classification F-measure (F-M) by up to 46\%.
We show our approach on the time series thermograms collected by our antenna testing system.
Conventionally, a variational autoencoder (VAE) learning observation noise may yield better results than a VAE with a constant noise assumption.
However, we demonstrate that this is not the case for anomaly detection on such low-level patterns for two reasons.
First, the baseline metric reconstruction probability, which incorporates the learned observation noise, fails to differentiate anomalous patterns.
Second, the area under the receiver operating characteristic (ROC) curve of a VAE with a lower observation noise assumption achieves 11.83\% higher than that of a VAE with learned noise.
\end{abstract}


\section{Introduction}

Antenna arrays are widely used in different wireless communications systems, such as mobile networks, radio broadcast, and satellites. 
By adjusting the power and phase of the input signal to each array element, a beam of radio waves can be steered in a particular direction in order to maximize the gain towards the target device. 
However, due to the complex configurations (i.e., the cross product of the input power, phase, and array element), and the individual occurrence of faulty elements, antenna arrays may result in anomalous radiation pattern deviating from the original design. 
Antenna testing aims at ensuring that the array performs as desired and further improves the stability in our communication devices.

In an indoor range antenna testing setting, near field radiation patterns, i.e., spatial distributions of the electromagnetic (EM) field, are measured by a scanning probe. 
The probe moves in a regular interval over the antenna under test (AUT) depending on the geometrical surfaces \cite{antenna_book}. 
With the progress made in machine learning, researchers have started to adopt neural networks to diagnose faulty antennas based on the deviation patterns \cite{FAULT_ANTENNA, failure_antenna_ML}.
However, the scanning process is time consuming and analyzing near-field measurements requires profound domain knowledge.
This makes the antenna testing process difficult to be integrated into a smart production line.

\begin{figure}[t]
    \centering
   \includegraphics[width=0.5\textwidth]{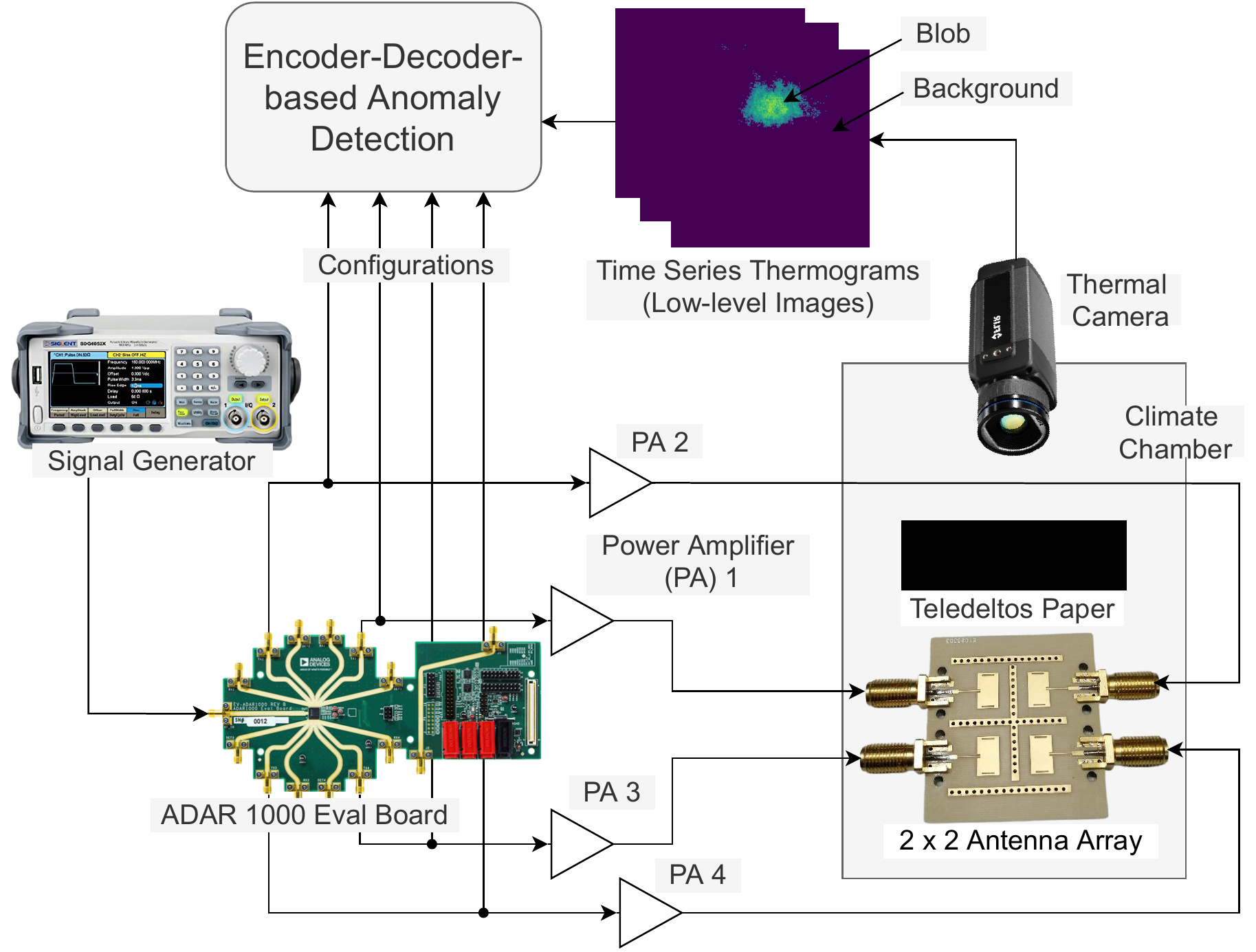}
    \caption{Schema of automated antenna testing system using time series thermograms. The condition denotes the power and phase of the input signal to  each array element. The time series thermograms contain blobs of various shape. Only the greenish blob pixels provide useful indications of anomalies, while the remaining pixels are background noise.}
    \label{fig:testing_setup}
\end{figure}

This paper uses an antenna testing system inspired by \citeauthor{Norgard} \cite{Norgard} and \citeauthor{thermography_pattern} \cite{thermography_pattern}, where a lossy material (e.g., Teledeltos paper) is positioned above the AUT to absorb the power of EM and infrared (IR) thermography is used to determine the magnitude of the EM field.
We propose to automate this system through a deep learning anomaly detector that tries to identify anomalous radiation patterns given the power and phase of the input signal that is fed into the individual elements. 
The automated antenna testing setup to detect faulty antenna arrays is depicted in Fig.~\ref{fig:testing_setup}. 

Our proposed solution uses a conditional variational autoencoder (VAE) composed of a convolutional neural network (CNN) and a long short term memory (LSTM)
to identify both contextual anomalies and collective anomalies \cite{anomaly_survey}. 
Contextual anomalies are instances whose behavior deviates from the rest of instances in the specific context.
A contextual anomaly occurs if the pattern in a thermogram does not conform to the normal pattern in the context of the input signal configuration.
Collective anomalies are a collection of instances that together express an anomalous behavior while individual instances themselves may be normal.
For instance, if pixel values in a normal thermogram remain the same over time although the expected behavior is a change, then a collective anomaly occurred.
Based on the proposed network architecture, we further apply four different variants to our anomaly detection and illustrate the peculiarity of image time series with low-level patterns.
We refer these images, which only contain blobs of various shapes and don't have a semantic representation, as 'low-level images'. 
These images have the characteristic that their reconstruction probability \cite{An2015VariationalAB} fails to alarm anomalies.
Finally, we manage to map the sparse reconstructed error to a decision by introducing a contour-based anomaly detector.

In summary, our contributions include:
\begin{itemize}
\item An automated antenna testing system that combines electromagnetic thermography with machine learning techniques.
\item A contour-based anomaly detection algorithm that can deal with sparse high dimensional data, as background noise would cancel out the residual error if a simple average were computed.
\item An analysis of encoder-decoder-based anomaly detection for high-dimensional time series data, in particular low-level images, which so far have limited coverage by literature.
\end{itemize}

\section{Related Work}

Since the temperature increase in the material is a function of time, \emph{time series anomaly detection} is used for our thermogram-based antenna testing task. 
The literature review has showed the focus on time series data usually falls into one of two categories: (multimodal) signal level and video level.

Researchers have made a great progress on the signal level. \citeauthor{Malhotra2015LongST} \cite{Malhotra2015LongST} introduced Long Short Term Memory (LSTM) to model prediction error distribution in space shuttle valve time series.
\citeauthor{ECG} \cite{ECG} utilized LSTM to analyze electrocardiograms. 
Since these prediction models may not detect anomalies in unpredictable time series, autoencoders (AEs) were introduced to overcome this problem using the reconstruction error as an anomaly score \cite{ecde_lstm,8260666}. 
As an alternative, variational autoencoders (VAEs) capture the underlying probability distribution of normal time series in anomaly detection tasks. 
Here, the anomaly score is measured through the reconstruction probability \cite{An2015VariationalAB}.
To detect anomalous task execution in robots, \citeauthor{park} \cite{park} presented an LSTM-based VAE using multimodal sensory signals. 
\citeauthor{VAE_BiLSTM} \cite{VAE_BiLSTM} took a different approach by applying a self-attention mechanism to VAE to improve the encoding-decoding process in energy data. 
Recently, \citeauthor{LSTMVAE_hybrid} \cite{LSTMVAE_hybrid} suggested a hybrid model, where the VAE learns short-term correlation and the LSTM learns long-term trends. 
However, our time series thermograms have higher dimensional space and even sparser information than those in multimodal signals: only the pattern pixels contain the useful indications of anomalies, while the remaining pixels are background noise.
Our approach overcome this issue using two methods.
First, a CNN is applied to the raw images, so that the LSTM learns from a more compact representation.
Second, the contour-based detector prevents the pattern pixels from being averaged out by the background noise.
 
With the breakthrough in computer vision, video anomaly detection has gained more attention. \citeauthor{DBLP:journals/corr/ChongT17} \cite{DBLP:journals/corr/ChongT17} presented a spatial AE coupled with temporal encoder-decoder for anomaly detection on surveillance footage.
\citeauthor{Chang2020ClusteringDD} \cite{Chang2020ClusteringDD} designed two AEs to distil the spatial and temporal information separately. \citeauthor{DBLP:journals/corr/abs-1805-11223} \cite{DBLP:journals/corr/abs-1805-11223} introduced a VAE framework to learn representations of normal samples as a Gaussian Mixture Model.
In industrial applications such as additive manufacturing, multimodal signals and videos are fused to calculate an anomaly score \cite{9356299}. 
Our data differs from these video datasets in the sense that thermograms contain less abstract information and features are less entangled.
We demonstrate that such properties of low-level images affect the performance of VAEs. 
To the best of our knowledge, this is the first time that a VAE has been considered for low-level image sequences.

In the application of antenna testing, existing AI-based approaches are based on supervised learning models \cite{FAULT_ANTENNA, failure_antenna_ML, Kaijing_array_FD,8971339}. 
A classifier is trained with normal data and anomalies to identify different types of anomalies, i.e., categorizing fault types rather than detecting anomalies. 
Hence, the models are incapable of classifying unseen instances that do not belong to any of the trained classes.
Knowing that modeling all types of anomalies is almost impossible, our proposed approach overcomes this problem through learning to model the normal behavior from the given training data. 

\section{Background}

\subsection{Infrared Thermography for Antenna Array Testing}

An antenna array is considered fault-free when it emits EM fields.
In our testing setup shown Fig.~\ref{fig:testing_setup}, the EM waves emanating from the antenna array are absorbed by Teledeltos paper placed above the antenna under test (AUT), resulting in a rise in temperature.
By measuring where and how much the temperature rises using a thermal camera, we are able to infer the shape and magnitude of the EM field, thereby identifying whether the antenna array is fault-free or not.
Since the Teledeltos paper can reach thermal equilibrium in the end, the absorbed power related to the magnitude of EM will equal the power lost due to thermal convection and thermal radiation. 
Thermal conduction is negligible since the thermal conductivity in the paper is low.
\citeauthor{Norgard} \cite{Norgard} proposed that a quadratic polynomial can fit this balance for the Teledeltos paper:

\begin{equation}
    k_1 E^2 + k_2 E + k_3 = T_s - T_a 
\end{equation}

where $E$ is the magnitude of EM field, $k_1$, $k_2$, and $k_3$ are constants depending on the system setup such as the sensitivity of the thermal camera, $T_s$ is the surface temperature, and $T_a$ is the ambient temperature. 
As the entire experiment is isolated from external disturbances using a climate chamber, $T_a$ can be considered as a constant as well.

\subsection{Autoencoder}
\label{sssec:Autoencoder}

An Autoencoder (AE) is a deep learning technique that learns a compact internal representation, the so-called latent variable, that best represents the input.
It consists of an encoder and a decoder network.
The encoder maps the input data $x \in \mathbb{R}^{d_{x}}$ to the latent space, while the decoder maps the latent variable $z \in \mathbb{R}^{d_{z}}$ back to the input space.
Normally, the latent space $\mathbb{R}^{d_{z}}$ has a lower dimensionality than the input space $\mathbb{R}^{d_{x}}$ (i.e., $d_{z} < d_{x}$), which means AEs learn a compressed representation of the input data.
During training, an AE tries to minimize the loss function that measures the difference between input $x$ and the reconstructed input $\hat{x}$ from the decoder.
Therefore, we are able to generate reconstructed input $\hat{x}$ that is as close as possible to input $x$.

In anomaly detection, AE-based approaches usually use the reconstruction error as an anomaly score.
An AE is trained only with normal data and learns how to reconstruct data with normal behaviors.
As for anomalies, their unforeseen anomalous behaviors will result in higher reconstruction error, since an AE never learns how to reconstruct data that lies outside of the normal data manifold.

\subsection{Variational Autoencoder}
\label{sssec:VariationalAutoencoder}

A Variational Autoencoder (VAE) \cite{kingma2014autoencoding} models the latent variable $z$ as random variable defined by the prior probability distribution $p(z)$, and hence can generate new samples that still lie inside the input data manifold.
The generation process is achieved by maximizing the marginal likelihood of training data.
For each individual data point $x$, we formulate\footnote{Note: $\theta$ and $\phi$ are the parameters of the encoder and the decoder, resp.} $p_{\theta}(x) = \int p_{\theta}(x | z)p(z)\,dz$.
To tackle this integral problem we can use Monte Carlo (MC) integration by sampling $z$ from the posterior $p_{\theta}(z | x)$.
However, the true posterior distribution is intractable for continuous latent space.
Using variational inference we are able to approximate $p_{\theta}(z | x)$ with $q_{\phi}(z | x)$ from a subset of a parameterized distribution.
The difference between the true posterior $q_{\phi}(z | x)$ and its approximation $q_{\phi}(z | x)$ can be measured by Kullback-Leibler divergence ($D_\text{KL}$):

\begin{equation}
\label{eqn:kl_post}
    \begin{aligned}
    D_{\text{KL}}(q_{\phi}(&z | x)\|p_{\theta}(z | x))=\log p_{\theta}(x) \\ 
    &-\mathbb{E}_{z \thicksim q} [\log p_{\theta}(x | z)]+ \KL{q_{\phi}(z | x)}{p(z)}
    \end{aligned}
\end{equation}

The second and third term of the right-hand side of Eq.~\ref{eqn:kl_post} are the negative Evidence Lower Bound (ELBO).
Minimizing $D_{\text{KL}}$ to find the best approximation in the distribution family parameterized by $\phi$ can be achieved by maximizing the ELBO, since the marginal log-likelihood $\log p_{\theta}(x)$ is independent of $\phi$.
On the other hand, ELBO is the lower bound of the marginal log-likelihood $\log p_{\theta}(x)$.
In order to maximize the marginal log-likelihood with respect to $\theta$ for the generation process, we maximize its ELBO. 
Combing these two optimization problems defines the loss function $\Loss_\text{VAE}$ for VAE:

\begin{equation}
\label{eqn:vae_loss}
    \Loss_{\text{VAE}} = \mathbb{E}_{z \thicksim q} [\log p_{\theta}(x | z)]
    - \KL{q_{\phi}(z | x)}{p(z)}
\end{equation}

The first term in the loss function is the expectation of reconstructing $x$ as realistic as possible given the latent variable $z$ sampling from $q_{\phi}(z | x)$.
The second term is the regularization of $z$ by minimizing the KL divergence between approximated posterior and our assumed prior $p(z)$.
Typically we assume the prior $p(z)$ follows a standard normal distribution $\mathcal{N}(0,1)$ and the approximated posterior $q_{\phi}(z | x)$ is a Gaussian distribution $\mathcal{N}(\mu_z,\sigma_{z}^2)$.
For our observation model $p_{\theta}(x | z)$, either a Bernoulli distribution or a Gaussian distribution can be chosen depending on the data.  
In our case, we choose a Gaussian distribution $\mathcal{N}(\hat{x},\sigma_{\hat{x}}^2)$ since temperature values in the thermograms are continuous data.
The log-likelihood term of a Gaussian distribution can be derived as Eq.~\ref{eqn:vae_loss}. Simply using mean squared error (MSE) to calculate the reconstruction loss is a special case of assuming a constant observation noise as 1 in the Gaussian observation model.

\begin{equation}
\label{eqn:variance_1}
    \log p_{\theta}(x | z) = -\frac{1}{2\sigma_{\hat{x}}^2}\norm{x-\hat{x}}^{2} - \log\sigma_{\hat{x}} \sqrt{2\pi}
\end{equation}

The choice of the observation noise $\sigma_{\hat{x}}^2$ is crucial, since it influences the local maxima corresponding to posterior collapse \cite{DBLP:journals/corr/abs-1911-02469}. 
Posterior collapse is an issue for VAE optimization where $q_{\phi}(z | x)$ collapses to a standard normal distribution, rendering the decoder unable to use the latent information anymore.
This would result in that the output of the decoder $\hat{x}$ is almost independent of $z$.
We investigate two types of VAEs to overcome this problem:

\paragraph{Probabilistic VAE} learns the observation noise $\sigma_{\hat{x}}^2$ together with the reconstructed mean $\hat{x}$ \cite{An2015VariationalAB}. 
The decoder architecture contains an additional head for estimating the noise given the latent variable $z$. 
In the case of images, we assume a diagonal covariance matrix for the image and let the network output one variance value per pixel and channel, meaning the outputted noise has the same dimensionality as the reconstructed mean.

\paragraph{$\beta$-VAE} is proposed by \citeauthor{Higgins2017betaVAELB} \cite{Higgins2017betaVAELB} by adding a weighting term on KL divergence.
The $\beta$ term can be seen as a constant observation noise that practitioners tune to circumvent the problem as shown in Eq.~\ref{eqn:vae_loss_beta}. 
Another way to interpret $\beta$ is that it balance the reconstruction and regularization terms.
$\beta > 1$ reduces the reconstruction quality and limits the representational capacity of $z$.
This forces the encoding process to capture the most salient features and further to keep conditionally independent features disentangled.
Disentanglement is useful for high-level computer vision tasks, considering we can simply control one dimension of $z$ to manipulate semantic features, such as generating faces with blue eyes.
For low-level images, it is the other way around: since features are less entangled, we are able to choose $\beta < 1$.

\begin{equation}
\begin{aligned}
\label{eqn:vae_loss_beta}
    \Loss_{\beta\text{VAE}}&= \mathbb{E}_{z \thicksim q} [\log p_{\theta}(x | z)]
    - \beta\KL{q_{\phi}(z | x)}{p(z)}\\
    & \sim -\frac{1}{2\beta}\norm{x-\hat{x}}^{2}
    - \KL{q_{\phi}(z | x)}{p(z)}
\end{aligned}
\end{equation}

\begin{figure*}[h!]
    \centering
    \includegraphics[scale=0.7]{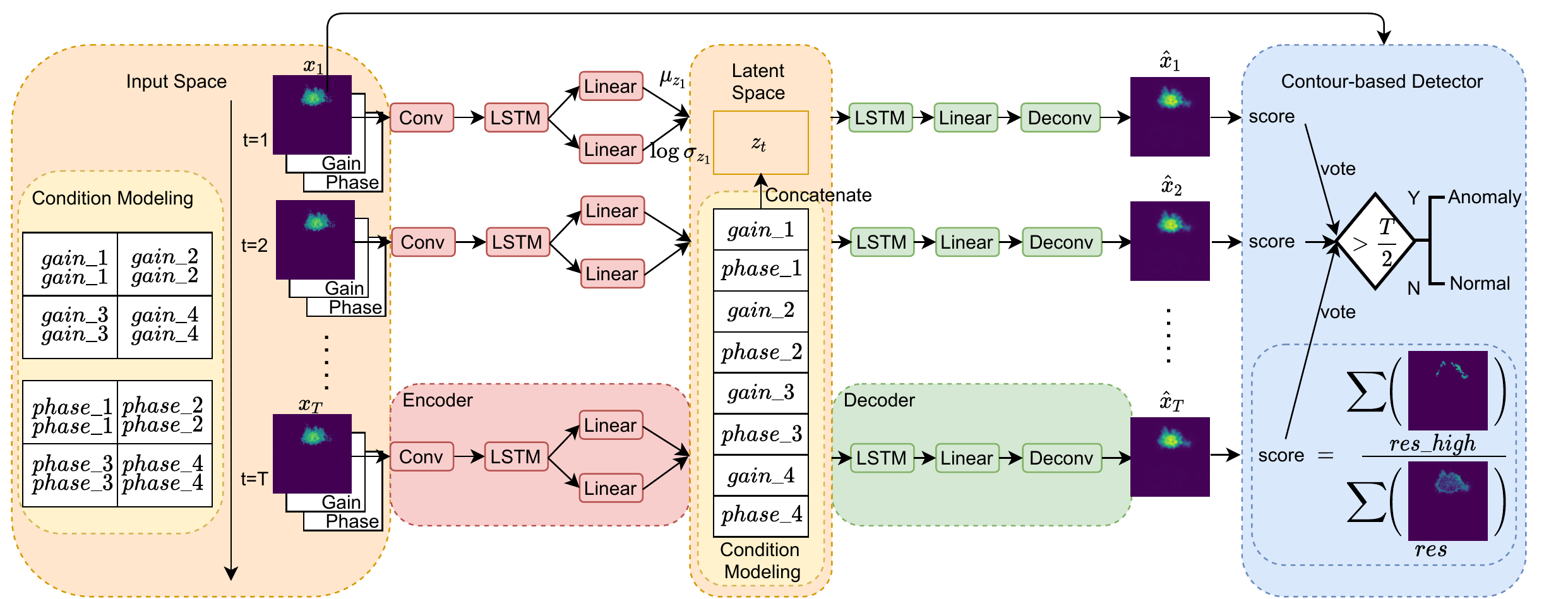}
    \caption{Unrolled conditional CNN-LSTM VAE with contour-based detector. The reconstructed $\hat{x}_t$ is the mean of observation model $p_{\theta}(x | z)$. Here a $\beta$-VAE model is presented to demonstrate the concept of time series thermograms anomaly detection. However this framework is applicable to probabilistic VAE and AE. For probabilistic VAE, decoder has another head to predict variance per pixel. For AE, the encoder requires only one linear head to learn a deterministic latent representation. The condition modeling modules are the same for all the models.}
    \label{fig:vae}
\end{figure*}

For anomaly detection, the reconstruction error in a VAE is defined by the difference between the input $x$ and the mean of observation model $\hat{x}$. 
Alternatively, a probabilistic VAE can measure anomalies by computing reconstruction probability, which takes the variability of the data into account.
The reconstruction probability is calculated as the MC estimation of $\mathbb{E}_{z \thicksim q} [\log p_{\theta}(x | z)]$.
The lower the reconstruction probability is, the higher chance the instance might be an anomaly. 
However, the challenge of learning the observation noise $\sigma_{\hat{x}}^2$ may induce reconstruction probability to fail as shown in Sec.~\ref{sec:evaluation} .

\section{Proposed Approach}

Our proposed anomaly detection framework consists of an encoder-decoder model and a contour-based detecting strategy.
The model takes multiple series of images as its input. 
Each sequence of images is denoted as $\series{x}=\{x_{1},x_{2},\dots,x_{T}\}$ where $T$ is the length of the image sequence.
Each image $x_{t}$ lies in a $d_x$-dimensional space, giving each image sequence a dimension of $(T,d_x)$.

\subsection{Conditional CNN-LSTM Variational Autoencoder}

To extract the spatial information and temporal correlation of the sequential input images, we propose a VAE composed of CNN and LSTM modules. 
We call the complete model CNN-LSTM VAE.
Fig.~\ref{fig:vae} shows the unrolled structure of the network.
In the encoder, the convolutional layers downsample the images.
The LSTM then aggregates the resulting features' local information, captures temporal dependency of the sequential data, and provides the cue for linear modules to estimate the global mean and variance of the approximated posterior $q_{\phi}(z_{t} | x_{t})$ for every time step $t$ of the image series.
The decoder will subsequently upsample a random sample $z_{t}$ from $q_{\phi}(z_{t} | x_{t})$ to get pixel-wise outputs.

We define the prior distribution $p_{\theta}(z)$ as a normal distribution $\mathcal{N}(z\lvert 0,1 )$.
One might argue for time-series data a static prior may degrade the approximated posterior $q_{\phi}(z_{t} | x_{t})$ to capture temporal dependency, because VAE tries to minimizes the difference between $q_{\phi}(z_{t} | x_{t})$ and $p_{\theta}(z)$.
\citeauthor{park} \cite{park} proposed a progress-based prior that gradually changes the prior mean using the mean of underlying distribution of inputs as the reference.
Since our sequential images are sparse, meaning the mean of the distribution of input images is almost 0 for every time step, we expect that using a standard normal distribution with a looser regularization term (smaller $\beta$) can still introduce temporal dependency to the approximated posterior.

To detect contextual anomalous patterns we need to model the power and phase of the input signal to each array element as a condition.
The power and phase configurations for an antenna array are denoted as $c = \{gain\_k, phase\_k\}_{k=1}^{K}$, where $gain\_1$ and $phase\_1$ represent the power and phase values of the first element in the antenna array.
Inspired by the conditional VAE \cite{cvae}, we model the condition in input space based on the spatial dependency of the element in the antenna array.
In the latent space, we design a vector composed of $gain\_k$ and $phase\_k$ to model the condition. 
Both condition maps are normalized in the range of [0,1]; one is then concatenated to the input data, and the other to the latent variable (see Fig.~\ref{fig:vae}).

\subsection{Contour-based Anomaly Detector}

The contour-based anomaly detector is built upon using the reconstruction error as an anomaly score metric. We calculate the absolute difference between ground truth (i.e., image taken during test) and reconstruction (i.e., model output).
This constitutes the reconstruction residual error for every time step $\series{r}=\{r_{1},r_{2},\dots,r_{T}\}$.

Alg.~\ref{alg:contour} shows the pseudo code for our contour-based anomaly detector taking the residual error as input. 
Firstly, we convert the residual error $r_{t} \in \mathbb{R}^{d_x}$ at time step $t$ to a binary image $binary\_1$ by a non-zero thresholding ($threshold=0$). 
A contour-finding algorithm suitable for binary images \cite{contour_paper} is then applied to $binary\_1$ to obtain the contour $contour\_1$ of the residual error. 
We then sum the pixels of residual error inside $contour\_1$ as $res$, representing the area of the contour.
Secondly, we want to scoop up pixels with greater residual error. 
To do so, residual error $r_{t}$ is converted to $binary\_2$ using a fixed-level $threshold$ which is computed by averaging the top $k$ pixel values of residual error ($k=5000$). 
We apply the contour-finding algorithm to $binary\_2$ and sum up the pixels inside $contour\_2$ to obtain the area of pixels with greater error $res\_high$. 
The quotient of $res\_high / res$ gives us an anomaly score for each time step.
Finally using a voting strategy, if more than half of the residuals in the sequence vote for anomalous, the image sequence is detected as an outlier; 
the antenna array is estimated as faulty.

\begin{algorithm}
  \caption{Contour Based Anomaly Detector}
  \label{alg:contour}
  \textbf{Input:} $\series{r}\in\mathbb{R}^{T\times d_x}$ \\
  \textbf{Output:} $Anomaly$ or $\neg Anomaly$
  \begin{algorithmic}[1]
  \State $vote \gets 0$
  \For{t = 1 to $T$}
        \State $binary\_1$ $\gets$ Convert to Binary($r_{t}$, 0)
        \State $contour\_1$ $\gets$ Find Contour($binary\_1$)
        \State $res$ $\gets$ Sum pixels inside $contour\_1$
        \State
        \State $threshold$ $\gets$ Average of top $k$ pixels
        \State $binary\_2$ $\gets$ Convert to Binary($r_{t}$, $threshold$)
        \State $contour\_2$ $\gets$ Find Contour($binary\_2$)
        \State $res\_high$ $\gets$ Sum pixels inside $contour\_2$
        \State $anomaly\_score$ $\gets$ $res\_high / res$
        \If{$anomaly\_score>\epsilon$}
            \State $vote$ $\gets$ $vote+1$
        \EndIf
  \EndFor
  \If{$vote > T/2$}
    \State \textbf{return} $Anomaly$
  \EndIf
  \end{algorithmic}
\end{algorithm}

The motivation of using contour to detect anomaly is that at each time step, $r_{t}$ lies in a high dimensional reconstruction space rather than a 1-D vector space (e.g., signals) as in most of the time-series anomaly detectors.
Mapping the residual error from $\mathbb{R}^{d_x}$ to a scalar $anomaly\_score$ becomes the core idea of designing the detector algorithm.
Simply averaging all the pixels will vanish the residual error since the residual is even sparser than our input.
\citeauthor{8999143} \cite{8999143} has introduced a spatial scoping window to find the maximum summation of the window from the residual error matrix. 
However, this approach has a problem in two aspects: 
(1) The size of the pattern varies and finding a suitable window is not simple.
(2) Patterns with larger sizes tend to become anomalies.
By summing pixel values inside the contour, we are able to focus on the residuals themselves instead of the background.
Furthermore, contour filters out small noise patches outside the main residual we are interested in. 
Thus using a contour-based detector has several advantages in the low-level image reconstruction error anomaly detection.

\section{Experiments and Dataset}


\subsection{Antenna Array Testing System Setup}

A signal generator is used to generate the frequency at 15.6 GHz and -4.3\si{\dBm} output power. This signal is fed into the ADAR 1000 Eval Board which splits the signal to the 4 ports. Each signal is fed into a power amplifier (PA) which amplifies to the required input power level. The output of the power amplifier is then fed to each antenna in the array.

\begin{figure}[h]
\centering
\begin{minipage}{.25\textwidth}
  \centering
  \includegraphics[height=3cm, width=0.9\linewidth]{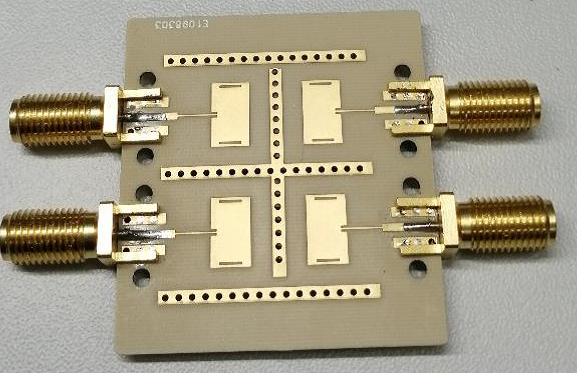}
  \captionof{figure}{2$\times$2 AUT}
  \label{fig:testing}
\end{minipage}%
\begin{minipage}{.25\textwidth}
  \centering
  \includegraphics[height=3cm, width=0.9\linewidth]{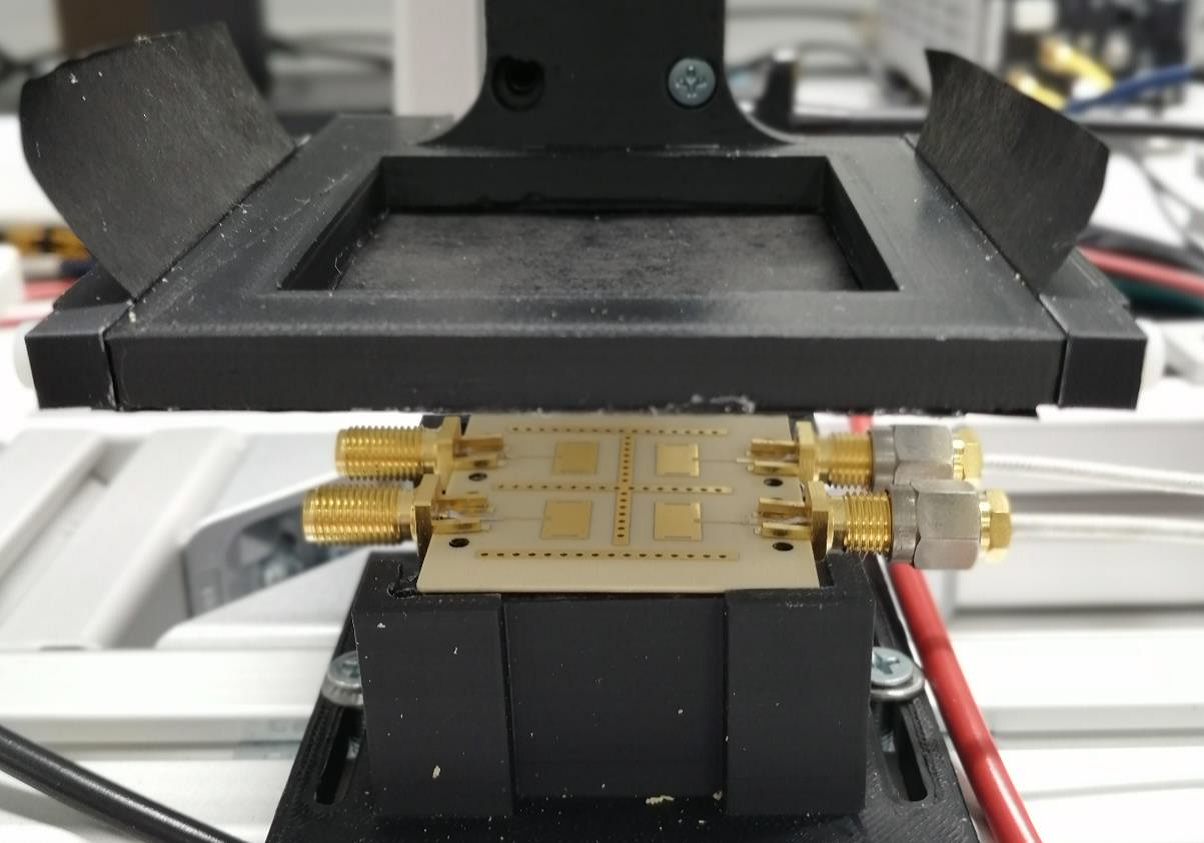}
  \captionof{figure}{Setup without insulation}
  \label{fig:test2}
\end{minipage}
\end{figure}

The AUT is a 2$\times$2 patch antenna array interfaced with 4 SubMiniature version A (SMA) connectors (see. Fig \ref{fig:testing}). Above the antenna array a Teledeltos paper is placed above the antenna at a distance of \SI{9.55}{\mm} which is the boresight distance for this configuration of the patch antenna. The dimensions of the paper are set to \SI[product-units=power]{50 x 50}{\mm} with a thickness of \SI{80}{\micro\metre} allowing to cover the whole aperture of the AUT and leaving some extra margins. A thermal camera, positioned at the focal distance of the lens, detects the change in temperature on the surface of the paper. The camera used in this experiment is from Flir A615 operating at room temperature. Since Flir A615 is a GenICam, any GenICam compliant Software with a driver could be used. In this experiment, we had used Stemmer Imaging GenICam Driver for Image acquisition and Halcon for processing the images from the Camera. This whole system is insulated against conduction and convection  heat loss by building inside a climate chamber.

\begin{figure*}[t]
\centering
    {\includegraphics[width=1.5cm]{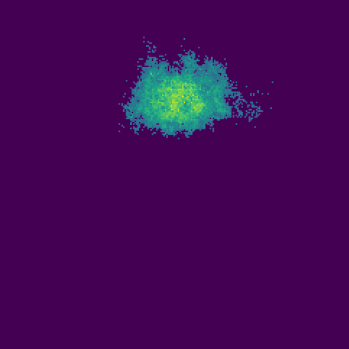}}\hspace{0.5em}\vspace{0.5em}%
    {\includegraphics[width=1.5cm]{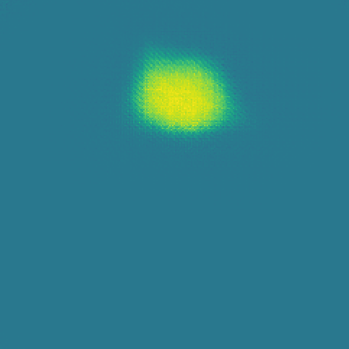}}\hspace{0.5em}%
    {\includegraphics[width=1.5cm]{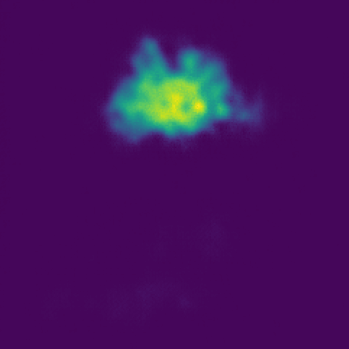}}\hspace{0.5em}%
    {\includegraphics[width=1.5cm]{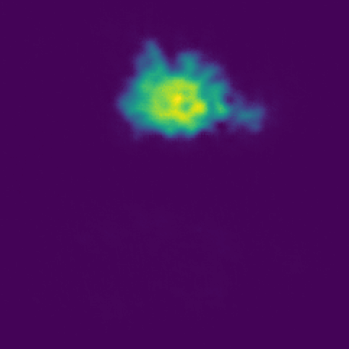}}\hspace{0.5em}%
    {\includegraphics[width=1.5cm]{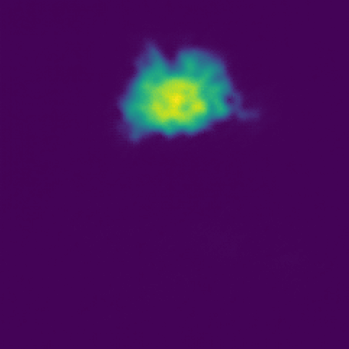}}
    {\includegraphics[width=1.5cm]{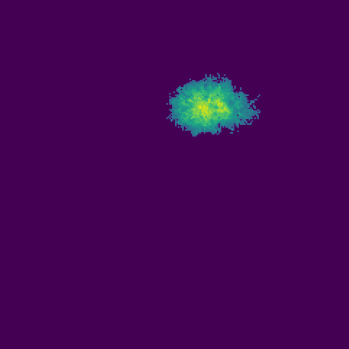}}\hspace{0.5em}\vspace{0.5em}%
    {\includegraphics[width=1.5cm]{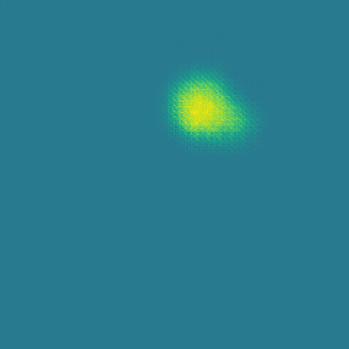}}\hspace{0.5em}%
    {\includegraphics[width=1.5cm]{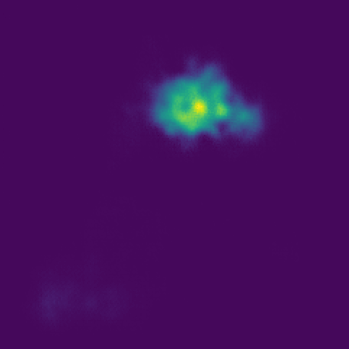}}\hspace{0.5em}%
    {\includegraphics[width=1.5cm]{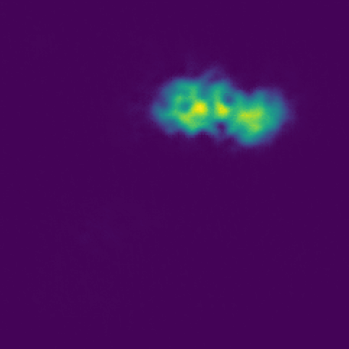}}\hspace{0.5em}%
    {\includegraphics[width=1.5cm]{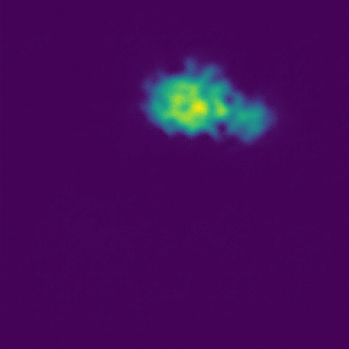}}\hspace{3em}%
    
    \subcaptionbox{GT\label{fig:gt_normal}}{\includegraphics[width=1.5cm]{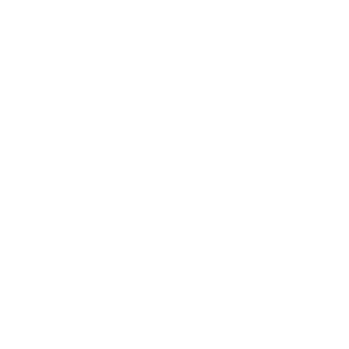}}\hspace{0.5em}\vspace{0.5em}%
     \subcaptionbox{PCVAE\label{fig:pcvae_normal}}{\includegraphics[width=1.5cm]{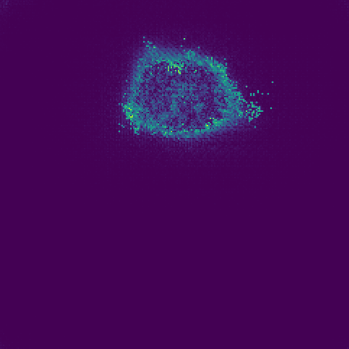}}\hspace{0.5em}%
     \subcaptionbox{CVAE}{\includegraphics[width=1.5cm]{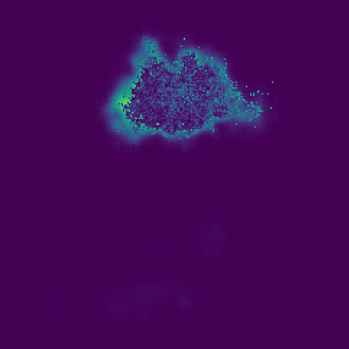}}\hspace{0.5em}%
     \subcaptionbox{0.01C.}{\includegraphics[width=1.5cm]{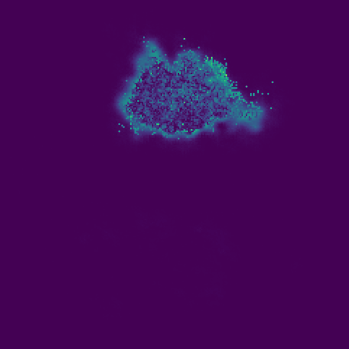}}\hspace{0.5em}%
     \subcaptionbox{AE}{\includegraphics[width=1.5cm]{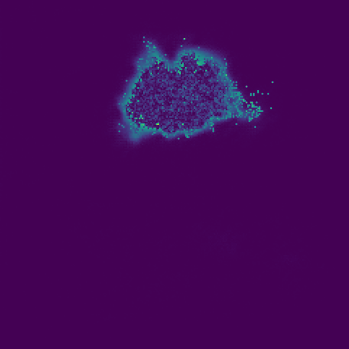}}\hspace{0.5em}%
    \subcaptionbox{GT\label{fig:gt_anomalous}}{\includegraphics[width=1.5cm]{fig/blank.png}}\hspace{0.5em}%
     \subcaptionbox{PCVAE\label{fig:pcvae_anomalous}}{\includegraphics[width=1.5cm]{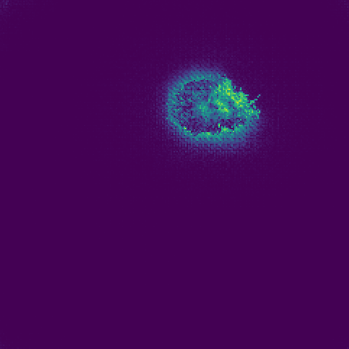}}\hspace{0.5em}%
     \subcaptionbox{CVAE}{\includegraphics[width=1.5cm]{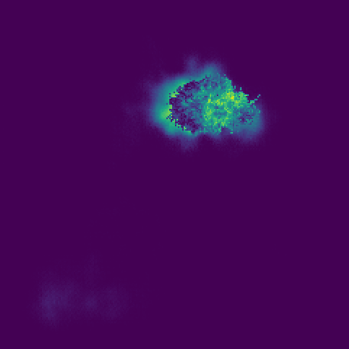}}\hspace{0.5em}%
     \subcaptionbox{0.01C.}{\includegraphics[width=1.5cm]{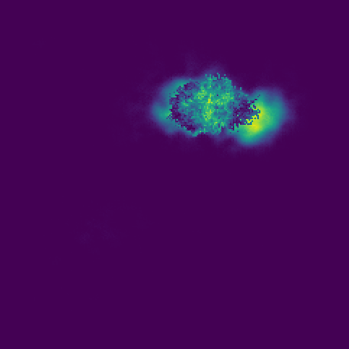}}\hspace{0.5em}%
     \subcaptionbox{AE}{\includegraphics[width=1.5cm]{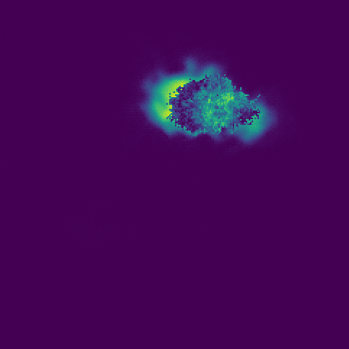}}\hspace{0.5em}%
    \caption{(a) Ground truth (GT) from a representative image of a normal sequence; (b-e) reconstructions (first row) and residuals (second row) for PCVAE, CVAE with observation SD of 1, CVAE with observation SD of 0.01, and AE, resp.; (f) ground truth from a representative image of an anomalous sequence with the same configuration and time step as (a); (g-j) reconstructions and residuals from anomalous data.}
    \label{fig:recon_result}
\end{figure*}

\subsection{Dataset}
\label{sssec:Dataset}

\begin{table}[]
\centering
\begin{tabular}{l|cccc}
Gain & Power\_1 & Power\_2 & Power\_3 & Power\_4 \\ \hline
255  & 22.81           & 22.41           & 22.00           & 22.81           \\
235  & 21.91           & 21.58           & 21.30           & 22.15           \\
185  & 16.03           & 15.85           & 15.65           & 17.03           \\
170  & 12.57           & 12.48           & 12.21           & 13.92           \\
160  & 9.40            & 9.38            & 8.99            & 11.09           \\
155  & 7.65            & 7.47            & 7.18            & 9.40 
\end{tabular}
\caption{ Relationship between the selected gain values and the powers (\si{\dBm}) of individual antenna in AUT}
\label{tab:gain_power_relation}
\end{table}

The data in this work is a set of sequential images acquired over a defined time interval by the thermal camera capturing different antenna configurations.
The power of the input signal to the antenna array is ranged from 7~\si{\dBm} to 22~\si{\dBm}.
In the ADAR board, we can set a gain value in steps of 5 from 0 to 255.
Based on the power with which the antenna is operated, we choose the gain values in the sense that the corresponding power levels differ between 2~\si{\dBm} to 6~\si{\dBm} as listed in Table~\ref{tab:gain_power_relation}.
We set the input signal’s phase values at every 45 degrees from 0 to 180.
Overall, we choose $\{ 155, 160, 170, 185, 235, 255 \}$ as our gain value set and $\{ 0, 45, 180, 135, 180 \}$ as our phase value set.
As the AUT is a two-by-two antenna array, each image sequence has a configuration of four gain-phase pairs.
For our dataset, we obtained 1,997 configuration samples randomly selected from the gain and phase value set.

The thermal images for each configuration are recorded over an interval of 100~\si{\ms} for 100 images per sequence.
The raw thermal images are stored as 16-bit grayscale TIFF.
The data is preprocessed by subtracting the background noise, which is the first image in the sequence, from the rest of the image and then normalized to the range of pixel values in the interval of [0, 1].
We divide our dataset of normalized sequences into training, validation, and test set with a splitting ratio of 80/10/10, respectively.
The anomalous data is created by placing an attenuator under AUT to synthesize faulty elements in the antenna array. 
Configurations used to generate anomalies correspond to those in our normal test set, and hence the normal and anomalous data in our test set are equally balanced. 

\subsection{Network Implementation}

We used TensorFlow \cite{tensorflow} to implement our encoder-decoder models.
A sliding window approach with length $T$ was set to 10 and offset set to 5 during training.
An 8-dimensional latent space was set for all models. 
The models were trained until the validation loss didn't decrease for 4 epochs. 

\section{Evaluation Results}
\label{sec:evaluation}
To evaluate the performance of our proposed approach, we implemented four encoder-decoder-based models with the same conditional CNN-LSTM structure:

\begin{itemize}
  \item Probabilistic conditional CNN-LSTM VAE (`PCVAE'): 
  PCVAE learns the observation noise from the model, and hence there is no need to set the $\beta$ value.
  \item Conditional CNN-LSTM VAE with $\beta = 1$ (`CVAE'): 
  The decoder learns to reconstruct the ground truth. By setting $\beta$ to 1, we assume that the observation model $\log p_{\theta}(x | z)$ is a normal distribution with standard deviation (SD) 1 according to Eq.~\ref{eqn:variance_1}.
  \item Conditional CNN-LSTM VAE w/ $\beta = e^{-4}$ (`0.01CVAE'): 
  Setting $\beta$ to $e^{-4}$ can be approximated  by assuming the SD of our observation model is 0.01.
  \item Conditional CNN-LSTM AE (`AE'): AE removes the stochastic parts (e.g., reparameterization trick and KL divergence) from the model. AE learns only a deterministic mapping from the input to the reconstruction space.
\end{itemize}

\subsection{Reconstruction Qualitative Results}

We first investigate the reconstruction quality of our proposed model for both normal sequence and anomalous sequence.
Fig.~\ref{fig:recon_result} shows the qualitative results for both the reconstructions from ground truth and the residuals.
Fig.~\ref{fig:recon_result}b-e show all models can reconstruct the normal pattern well.
Fig.~\ref{fig:gt_anomalous} has an anomalous pattern in the thermal image, which is difficult to discriminate from normal patterns without knowing the configuration setting.
Interestingly, Fig.~\ref{fig:recon_result}g-j show that when the model is less probabilistic, the reconstruction deviates more from the ground truth.
This is due to the fact that the anomalous data does not differ much from the normal data.
A probabilistic model with high variances such as PCVAE and CVAE can still fit anomalous data into its observation model $\log p_{\theta}(x | z)$.
Therefore, these models are insensitive to the anomalous data.
On the other hand, AE learns only a deterministic mapping to reconstruct the data, making the model sensitive enough to perceive the difference in the anomalous data.
An interesting observation is that 0.01CVAE not only preserves the sensitivity similar to AE, but is also able to tolerate some variation from the data (e.g., thermal patterns may be slightly different in other experiments due to aleatory variability of our setup).
Furthermore, the latent space is less constrained with a lower value of $\beta$.
This result is consistent with what we described in Section~\ref{sssec:VariationalAutoencoder} for low-level images, which do not have highly abstract features that need to be disentangled.

\subsection{Anomaly Score Analysis}

\begin{figure}[t]
    \centering
    \subcaptionbox{normal log-variance}{\includegraphics[width=0.4\linewidth]{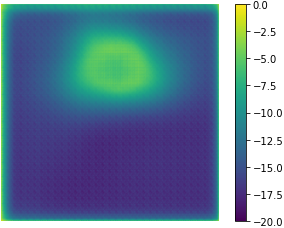}}\hspace{0.8em}%
    \subcaptionbox{anomalous log-variance}{\includegraphics[width=0.4\linewidth]{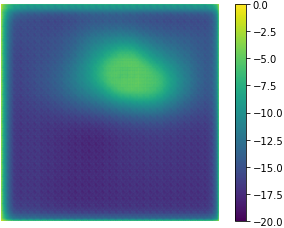}}\hspace{0.5em}
    \caption{Visualization of log-variance for both normal pattern and anomalous pattern using PCVAE from Figs. \ref{fig:gt_normal} and \ref{fig:gt_anomalous}}
    \label{fig:logvar}
\end{figure}

\textbf{Limitation of Reconstruction Probability: } Reconstruction probability is a baseline method using PCVAE for anomaly detection.
The probability is computed as the likelihood of the input data given the latent variable.
However, this method fails to detect our anomalous data for two reasons.
First, reconstruction probability assumes anomalous samples have higher variability than normal data, and thus gives a higher anomaly score for data with high variance.
However, the anomalous patterns do not appear to have a higher variance than normal patterns as can be seen in Fig.~\ref{fig:logvar}.
Secondly, the model assigns the location of the pattern with higher variance.
Therefore, residuals in these locations will be tolerated.
As a result, the reconstruction assigned with a higher probability is not due to a smaller residual error in the pattern, but rather due to the smaller size of the pattern.
An input image without any pattern but only background may be considered as a normal pattern, which is not true.  
Due to these two reasons, the contour-based detector is built upon reconstruction error rather than reconstruction probability. 

\begin{table}[t]
\centering
\begin{tabular}{ll|rrr}
 &Methods & Sn(\%) & Pr(\%) & F-M(\%)\\ \hline
\multirow{4}{*}{\RNum{1}} &PCVAE  & 75.0           & 76.9           & 75.9                     \\
&CVAE  & 83.6           & 92.5           & 87.8                      \\
&0.01CVAE  & 91.1           & 92.5           & \textbf{91.8}                  \\
&AE  & 89.6           & 90.5           & 90.0                         \\ \hline
\multirow{4}{*}{\RNum{2}} &PCVAE (w/o voting)  & 69.0            & 80.9            & 74.4    \\       
&CVAE (w/o voting)   & 71.9           & 91.4            & 80.5\\
&0.01CVAE (w/o voting)   & 85.4           & 93.4            & \textbf{89.2}\\ 
&AE (w/o voting)   & 84.8           & 92.0            & 88.3\\ \hline
\multirow{4}{*}{\RNum{3}} &PCVAE (w/o contour)  & 29.0            & 30.7            & 29.8    \\       
& CVAE (w/o contour)   & 44.0           & 60.3            & 50.8\\
&0.01CVAE (w/o contour)   & 48.1           & 44.2            & 46.1\\ 
&AE (w/o contour)   & 46.2           & 42.7            & 44.4\\ 
\end{tabular}
\caption{Evaluation results of F-measure (F-M), sensitivity (Sn) and precision (Pr) for contour-based anomaly detector}
\label{tab:qualitative}
\end{table}

Quantitative evaluations are shown in Table~\ref{tab:qualitative}, where we evaluate four implemented models combined with our contour-based detector.
We also perform two ablation studies: in Row~\RNum{2} the voting loop is removed and an anomaly is decided by the score $res\_high / res$ directly and in Row~\RNum{3} the contour anomaly score is replaced with the mean of residuals.
We use the following quality metric to evaluate the results \cite{IntroML}:
Sensitivity ($Sn=\frac{TP}{TP+FN}$) measures the fraction of how many are correctly detected among all anomalous data; Precision ($Pr=\frac{TP}{TP+FP}$) provides the ratio of how many are correctly detected out of all detected as anomalies; F-Measure ($F-M=2\cdot\frac{Pr\cdot Sn}{Pr+Sn}$) balances between $Sn$ and $Pr$.
The 0.01CVAE contour-based anomaly detector outperforms other models with all measurements on our test set.
In contrast to Row~\RNum{1} which evaluates the classification scores in a sequence level, Row~\RNum{2} evaluates those in a per image level.
By removing the voting procedure, we can further observe whether the classification results of the sequence is consistent with the images inside itself, e.g., a sequence is classified as normal, but a quarter of the images are detected as outliers, which indicates a high uncertainty classification in the sequence level. 
By comparing Rows \RNum{1} and \RNum{3}, it is clear that our contour-based detector improves the performance for all models;
Simply summing up background errors will degrade anomaly detection performance.

\begin{figure}[t]
    \subcaptionbox{PCVAE\label{fig:PCVAE_score}}{\includegraphics[width=0.49\linewidth]{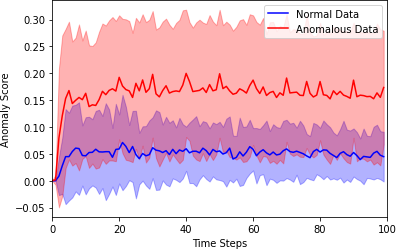}}\hspace{0.3em}%
    \subcaptionbox{CVAE\label{fig:CVAE_score}}{\includegraphics[width=0.49\linewidth]{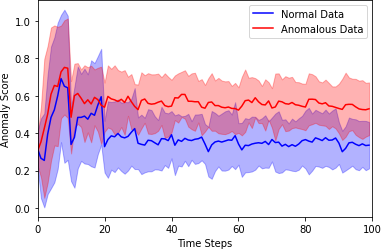}}
    
    \subcaptionbox{0.01CVAE\label{fig:0.01CVAE_score}}{\includegraphics[width=0.49\linewidth]{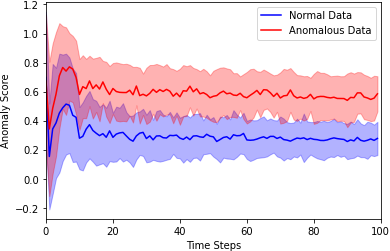}}\hspace{0.3em}%
    \subcaptionbox{AE\label{fig:AE_score}}{\includegraphics[width=0.49\linewidth]{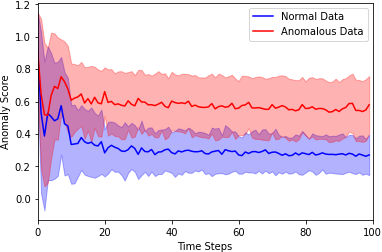}}
    \caption{Example distribution of anomaly scores over time from 32 normal thermal image sequence and 32 anomalous sequence with the same configuration}
    \label{fig:score}
\end{figure}

\begin{figure}[t!]
    \centering
    \includegraphics[width=0.5\textwidth]{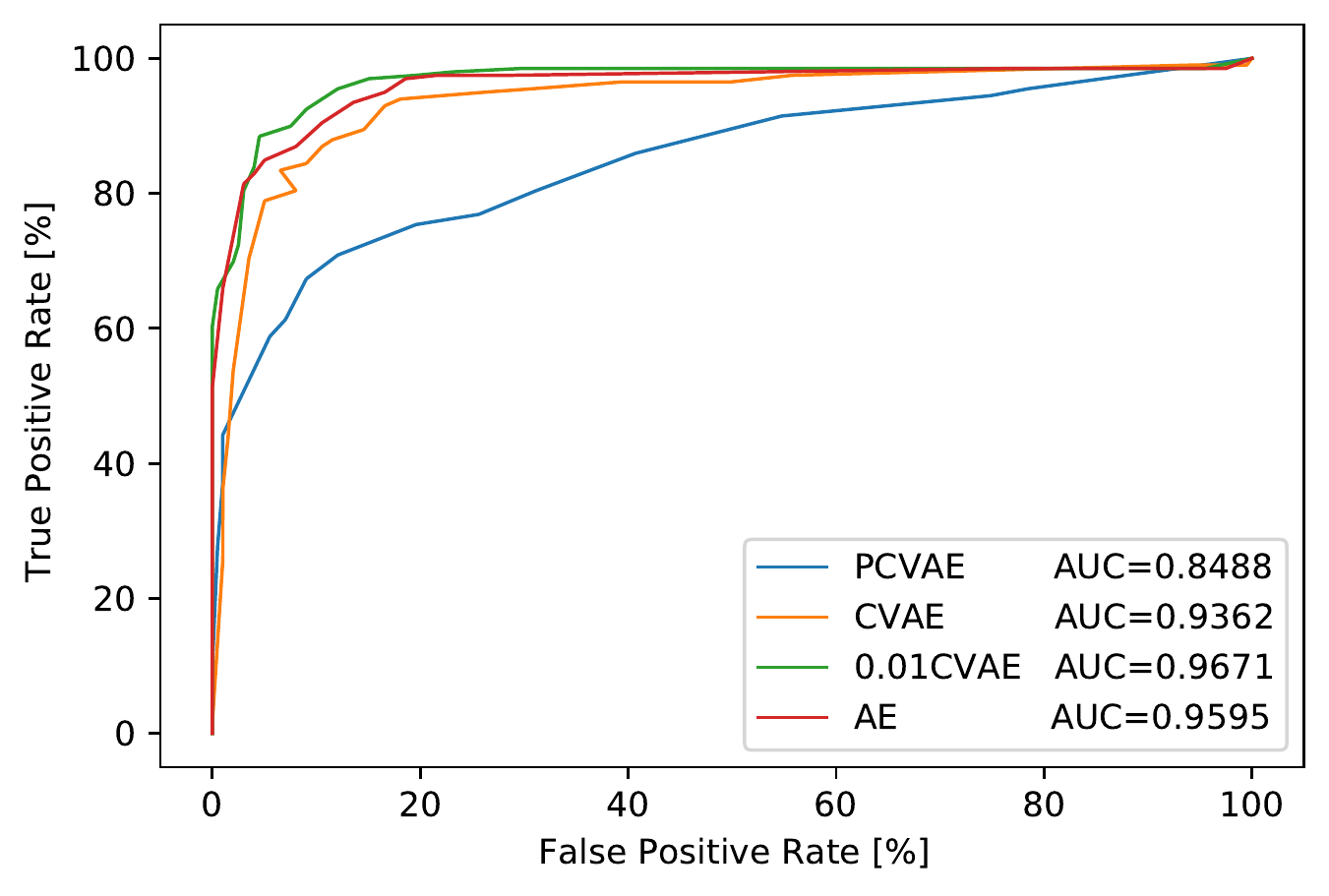}
    \caption{Receiver operating characteristic (ROC) curves over the entire test set}
    \label{fig:ROC}
\end{figure}

Fig.~\ref{fig:score} shows the distribution of anomaly scores over time from the contour-based detector testing with 32 normal image sequences and 32 anomalous sequences with the same configuration.
The blue and red shaded regions show the mean and SD of anomaly scores of normal and anomalous data, respectively.
Note that at the beginning of the time steps, the Teledeltos paper just starts to absorb the radiation emitted by AUT and therefore has a higher SD.
As the Teledeltos paper keeps absorbing power, the pattern in the thermal image becomes clearer, allowing our model to ensure whether it is an anomaly or not through a lower SD of the anomaly scores.
When there is less overlap between the score of normal and anomalous data, the anomaly score is effective enough to distinguish anomalies.
Thus, Fig.~\ref{fig:0.01CVAE_score} visualizes the advantage of our proposed approach with the 0.01CVAE.

Receiver operating characteristic (ROC) curves from four models are shown in Fig.~\ref{fig:ROC}. 
We show that 0.01 CVAE also outperforms other models with a higher area under the ROC curve (AUC). 

\section{Conclusion}

In this paper, we propose an automated antenna testing system by combining the concept of measuring EM using thermograms and machine learning techniques.
Within that, we investigate time series anomaly detection for low-level image sequences, where different encoder-decoder-based models are analyzed in the context of variance modeling.
We further introduce a contour-based detector to overcome the issue of background noise in sparse images.
Finally, we demonstrate that learning the observation noise from neural network does not show better performance.  
The AUC of a lower variance assumption VAE performs 11.83\% higher than that of a variance-learning-VAE for low-level image anomaly detection tasks.
We hope this paper can trigger other investigations into low-level image sequences anomaly detection, i.e., on image sequences that only contain low-level patterns such as blobs. 

{\footnotesize
\bibliography{bibliography}}
\bibliographystyle{IEEEtranN}

\end{document}